\documentclass[runningheads]{llncs}

\usepackage{eccv}

\usepackage{eccvabbrv}

\usepackage{graphicx}
\usepackage{booktabs}

\usepackage[accsupp]{axessibility}  
\usepackage{url}            
\usepackage{booktabs}       
\usepackage{amsfonts}       
\usepackage{nicefrac}       
\usepackage{microtype}      
\usepackage{xcolor}  
\usepackage{graphicx}
\usepackage{booktabs}
\usepackage{multirow} 
\usepackage{makecell}
\usepackage{ulem}
\usepackage[linesnumbered,ruled,vlined]{algorithm2e}
\SetKwRepeat{Do}{do}{while}%
\usepackage{amsmath} 
\usepackage{amssymb}
\usepackage{color, colortbl}
\definecolor{Gray}{gray}{0.9}
\usepackage{makecell}
\usepackage{wrapfig}

\usepackage{hyperref}

\usepackage{orcidlink}

\begin{document}

\title{MobileSAM2: Lightweight Segment Anything for Spatial Intelligence}

\titlerunning{MobileSAM2}

\author{
Kai Jiang\textsuperscript{1}\orcidlink{0000-0001-9921-2043} 
\and Jiaxing Huang\textsuperscript{\rm 1,}\thanks{Corresponding author.} \orcidlink{0000-0002-8681-0471} 
\and
Jingyi Zhang\inst{2}
\and Weiying Xie\inst{3}
\and Yunsong Li\inst{3}
\and Yufei Wang\inst{4}
\and Aoran Xiao\inst{2}
\and Dacheng Tao\inst{2}
}

\authorrunning{K.~Jiang et al.}

\institute{Hong Kong Polytechnic University, HK, China
\and Nanyang Technological University, Singapore \and
Xidian University, China
\and
SparcAI Inc., USA
\\
\email{xdjiangkai@foxmail.com, jiaxing.huang@polyu.edu.hk}
}

\maketitle

\begin{abstract}
The recent large video foundation model, SAM2, enables segment anything in both images and videos, serving as a powerful base model for various applications.
However, many of such use cases require to operate on resource-constrained devices like mobile phones and laptops.
In this work, we aim to make SAM2 more mobile-friendly by distilling the heavyweight SAM2 into a lightweight model, facilitating segment anything in both images and videos on mobile devices.
To this end, we propose Hypergraphical Knowledge Distill (HyperKD), which introduces the idea of hypergraph into knowledge distillation, aiming to effectively model and transfer SAM2's generalizable and comprehensive knowledge.
HyperKD consists of Temporal HyperKD and Granularity HyperKD that construct hypergraphs to explicitly model and extract the generalizable temporal knowledge and the comprehensive multi-granularity knowledge from SAM2 respectively, which are then distilled into the lightweight student model by aligning it with the constructed hypergraphs.
Besides, we present MobileSAM2, a new family of lightweight SAM2 that balances efficiency and effectiveness via searching the best model architectures with HyperKD during model size reduction.
Extensive experiments validate MobileSAM2 across multiple benchmarks and show promising generalization performance on embodied AI tasks.
  \keywords{Segment anything model \and Knowledge distillation \and Lightweight model \and Spatial intelligence \and Embodied AI}
\end{abstract}

\section{Introduction}
Large vision foundation models~\cite{radford2021learning,huang2024open,kirillov2023segment,ravi2024sam2} trained with large-scale visual data have achieved significant success across various areas of computer vision.
A prominent example is Segment Anything Model (SAM)~\cite{kirillov2023segment}, the first foundation model for image segmentation, which learns over 11 million images with 1.1 billion mask annotations and thus enables segmenting anything.
Recently, SAM2~\cite{ravi2024sam2} has extended SAM to video segmentation while keeping the image segmentation ability intact.
SAM2 learns rich temporal and segmentation knowledge from an unprecedented dataset with 113.8K videos and 40.9M mask annotations (e.g., SA-V manual and internal data~\cite{ravi2024sam2}) and outperforms traditional methods by substantial margins, showing great flexibility in segmenting anything in both still images and video streams and demonstrating strong generalization and zero-shot 

Given its impressive capabilities, SAM2 has been employed as a powerful foundation model for a wide range of computer vision applications, such as video editing~\cite{ceylan2023pix2video,liu2024video}, traffic surveillance~\cite{zhang2022monocular,zhang2020traffic}, autonomous vehicles~\cite{faisal2019understanding,parekh2022review}, robotics~\cite{dupont2021decade,javaid2021substantial}, etc.
On the other hand, many of these use cases require to operate on resource-constrained devices~\cite{shuvo2022efficient,kamath2023deep,kornaros2022hardware} like mobile phones, laptops, drones, robots, etc.
In this work, we aim to make SAM2 more mobile-friendly by distilling it into a lightweight model, facilitating segmenting anything on mobile devices for both images and videos.

\begin{figure}[t]
    \centering
    \includegraphics[width=0.65\textwidth]{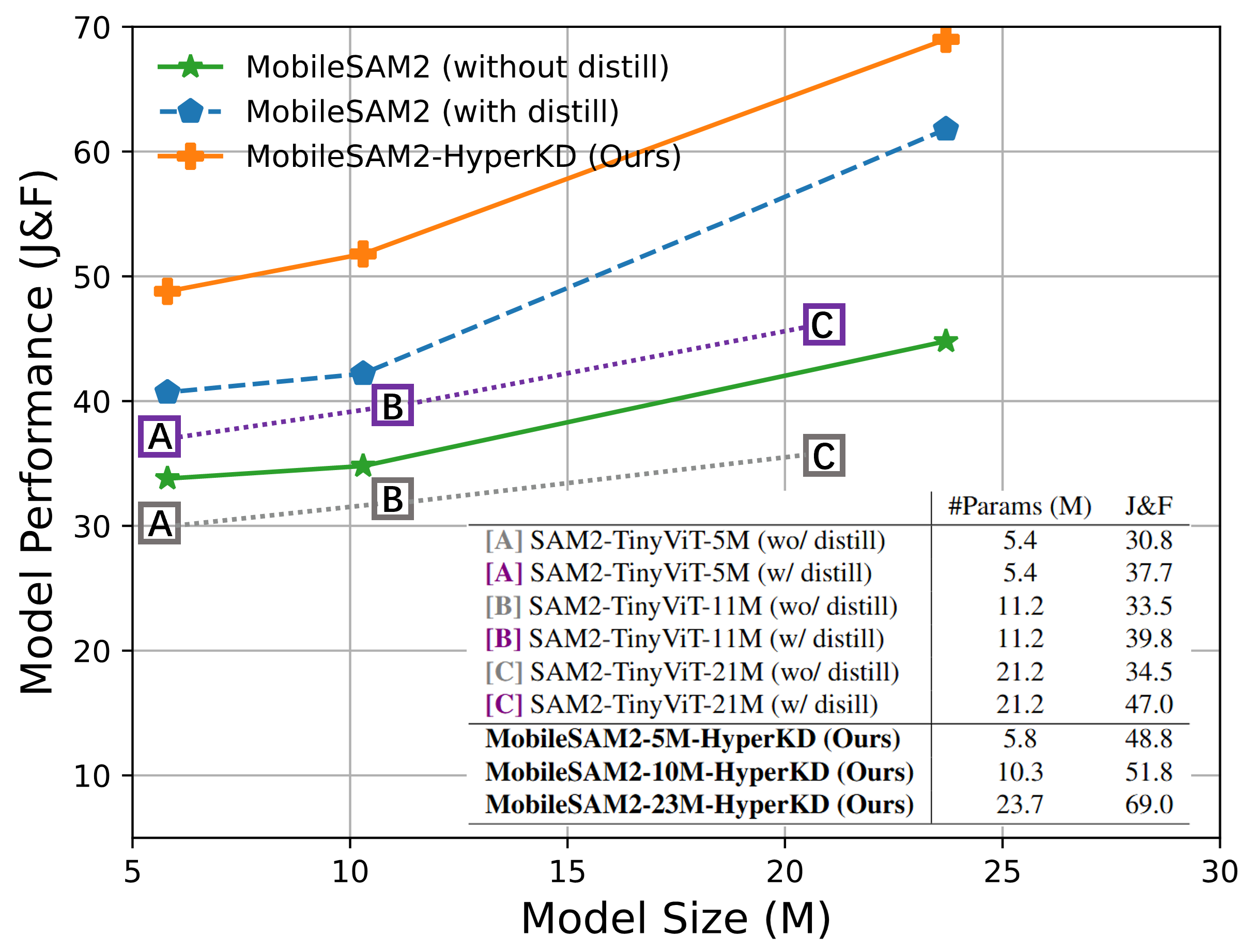}
    \vspace{-0.5em}
    \caption[Caption for LOF]{\textbf{Video Segmentation Comparison\protect\footnotemark.}
    With the proposed HyperKD and the searched model architectures, our MobileSAM2 works effectively with limited parameters.}
    \vspace{-2.5em}
    \label{fig:acc_para}
\end{figure}

\footnotetext{SAM2-TinyViT denotes SAM2 using TinyViT~\cite{wu2022tinyvit} as backbones. Please refer to Table~\ref{tab:result} for experiment details.}

We begin by examining the key factors behind SAM2's success.
Compared to previous video segmentation methods, SAM2, for the first time, scales up video training data to an unprecedented level, both in the number of videos and the granularity of mask annotations, learning from which two critical types of knowledge including
(1) Temporal Knowledge: SAM2 captures rich temporal correlations across video frames by learning from over 100K videos ($\sim$50 times of the previous largest dataset), ultimately acquiring generalizable temporal knowledge that enables robust segmentation along video frames;
and (2) Multi-Granularity Knowledge: SAM2 captures diverse granularity correlations across multiple granularity levels of visual concepts from over 40M multi-level mask annotations (e.g, objects as well as parts and subparts), finally learning comprehensive multi-granularity knowledge that enables comprehensive and coarse-to-fine understanding and segmentation of various visual concepts in videos.
Inspired by these insights, we believe that one effective method of distilling SAM2’s knowledge is to transfer such two types of knowledge comprehensively, such that the lightweight student model can inherit SAM2's robust and comprehensive video segmentation capabilities.

To this end, we propose Hypergraphical Knowledge Distill (HyperKD), which introduces the idea of hypergraph into knowledge distillation, aiming to effectively model and transfer SAM2's generalizable and comprehensive knowledge into a lightweight model.
Since these two types of knowledge are implicitly encoded in SAM2's learnt parameters and thus difficult to directly access, HyperKD constructs hypergraphs to explicitly model and extract the generalizable temporal knowledge and the comprehensive multi-granularity knowledge from SAM2, which are then distilled into the lightweight student model by aligning it with the constructed hypergraphs.

We instantiate HyperKD with two examples, named Temporal HyperKD and Granularity HyperKD.  
Temporal HyperKD first considers patches/objects within a single video frame as nodes and the distances between their embeddings as edges, and then constructs a hypergraph by linking multiple related nodes across frames via hyperedges that are formed by fusing edges from each individual frame.
Granularity HyperKD first treats segmentation entities within a single segmentation granularity level as nodes and the distances between their embeddings as edges, and then builds a hypergraph by linking multiple related nodes across segmentation granularity levels (e.g., objects, parts and subparts) via hyperedges that are formed by fusing edges from each individual segmentation granularity level.

Besides, based on our proposed HyperKD, we introduce MobileSAM2, a new family of lightweight segment anything models, which strikes a favorable trade-off between model size and performance.
Specifically, we adopt a progressive model contraction strategy~\cite{ryali2023hiera} to iteratively scale down a large model, generating a series of lightweight segment anything models by searching and identifying model architectures that best retain SAM2's temporal and multi-granularity knowledge (i.e., those can minimize HyperKD losses) during the model contraction process.

Our MobileSAM2 offers three desirable features:
(1) Robust Temporal Segmentation: Temporal HyperKD models and transfers SAM2’s temporal knowledge, enabling MobileSAM2 to capture rich correlations across video frames and achieve robust segmentation along video frames;
(2) Comprehensive Multi-Granularity Segmentation: Granularity HyperKD enables MobileSAM2 to capture diverse correlations across multiple segmentation granularity levels, allowing for comprehensive and coarse-to-fine understanding and segmentation;
(3) Efficiency\&Effectiveness: By searching for the best model architectures via HyperKD, the resulting MobileSAM2 strikes an effective balance between model size and performance.

The main contributions of this work are threefold.
First, we introduce the idea of hypergraph for knowledge distillation to explicitly model and transfer SAM2's generalizable and comprehensive knowledge into a lightweight model.
To our knowledge, this is the first work that explores hypergraph for video segmentation knowledge distillation.
Second, we design HyperKD, including Temporal HyperKD and Granularity HyperKD that distill temporal and multi-granularity knowledge effectively.
Third, we present MobileSAM2, a new family of lightweight segment anything models that effectively balances model size and performance.
Fourth, extensive experiments demonstrate MobileSAM2’s superior performance across multiple benchmarks.

\section{Related Works}
\subsection{Segment Anything Model}
Segmentation Anything Models (SAMs) have shown impressive zero-shot generalization and interactive segmentation across diverse datasets. Models like SAM~\cite{kirillov2023segment}, SEEM~\cite{zou2023segment}, and Semantic-SAM~\cite{li2023semantic} advance segmentation by using a promptable architecture that processes spatial (points, boxes, masks) and/or semantic (text) prompts to generate segmentation masks. Fast-SAM~\cite{zhao2023fast}, MobileSAM~\cite{zhang2023faster}, and TinySAM~\cite{shu2023tinysam} enhance inference speed, while HQ-SAM~\cite{ke2024segment} improves segmentation quality.
SAM2~\cite{ravi2024sam2} extends SAM~\cite{kirillov2023segment} to video segmentation with a memory-based transformer, enabling robust and comprehensive segmentation across video frames by storing object information and handling past interactions. 
However, SAM2’s large parameters and high computational demands make deployment on resource-constrained devices challenging.
Compressing and accelerating SAM2 is a largely under-explored problem. 
We focus on distilling SAM2 into a lightweight model, enabling efficient segmentation on mobile devices for both images and videos.

\subsection{Knowledge Distillation}
Knowledge distillation (KD)~\cite{gou2021knowledge} aims to train compact and efficient student models under the guidance of large teacher models. 
By learning from the teacher's soft labels, the student can outperform models trained only on hard labels. 
KD methods are broadly categorized into logit-based, feature-based, and relation-based approaches. 
Logit-based KD~\cite{chen2020online,hinton2015distilling,jin2023multi,mirzadeh2020improved, zhang2023not, zhang2018deep} minimizes a distance metric, such as KL divergence, between the predicted probabilities of the student and teacher, typically derived from the softmax of logit outputs. 
Feature-based KD~\cite{chen2022knowledge, chen2021distilling, guo2023class, heo2019comprehensive, yim2017gift, tian2022learning, tian2019contrastive, romero2014fitnets, liu2019structured, lin2022knowledge, li2021online} uses intermediate feature maps or refined information as knowledge. 
Relation-based KD~\cite{li2022knowledge, mei2023conditional, park2019relational, peng2019correlation, tung2019similarity, yang2021knowledge} aligns instance correlations between the student and teacher networks.
In this work, we aim to distill SAM2's generalizable temporal and multi-granularity knowledge learnt from over 100K videos and 40M annotations by capturing and encapsulating coarse-to-fine semantic relationships across frames, achieving robust and comprehensive video segmentation in a lightweight MobileSAM2.


\subsection{Hypergraph}
Hypergraph~\cite{feng2019hypergraph,jiang2019dynamic,bai2021hypergraph,feng2025hypergraph,gao2020hypergraph,wang2024graphs,lee2024survey,kim2024survey,gao2020hypergraph,gao2022hgnn+,feng2024hyper} is an advanced data structure that captures complex, high-order associations by allowing hyperedges to connect multiple nodes simultaneously. 
This capability makes hypergraphs particularly effective for modeling intricate relationships that traditional pairwise graph structures cannot adequately represent. 
Hypergraphs have been widely applied across domains such as social network analysis~\cite{young2021hypergraph,yang2020lbsn2vec++}, drug-target interaction modeling~\cite{jin2023general,vinas2023hypergraph}, and brain network analysis~\cite{xiao2019multi,zu2016identifying}, where multi-way interactions are crucial for understanding underlying structures.
In this work, we design a hypergraphical knowledge distillation method that constructs temporal and granularity hypergraphs to distill SAM2's generalizable knowledge learnt from a vast dataset. 
This approach effectively uncovers coarse-to-fine, high-order semantic relationships of various visual concepts across video frames, resulting in a lightweight MobileSAM2 with robust and comprehensive video segmentation capabilities.

\section{Methodology}
This section presents MobileSAM2, a new family of tiny and efficient segment anything models with effective and efficient distillation on video segmentation. 
We first introduce preliminaries in Section~\ref{sec: Preliminaries}.
Then we introduce the Hypergraphical Knowledge Distillation (HyperKD) framework for small segment anything model training in Section~\ref{sec: Hypergraphical Knowledge Distillation}. After that, we design a new tiny model family that balances efficiency and effectiveness by progressively scaling down a large seed model in Section~\ref{sec: Model Architectures}.

\begin{figure*}[t]
\centering
\includegraphics[width=1\linewidth]{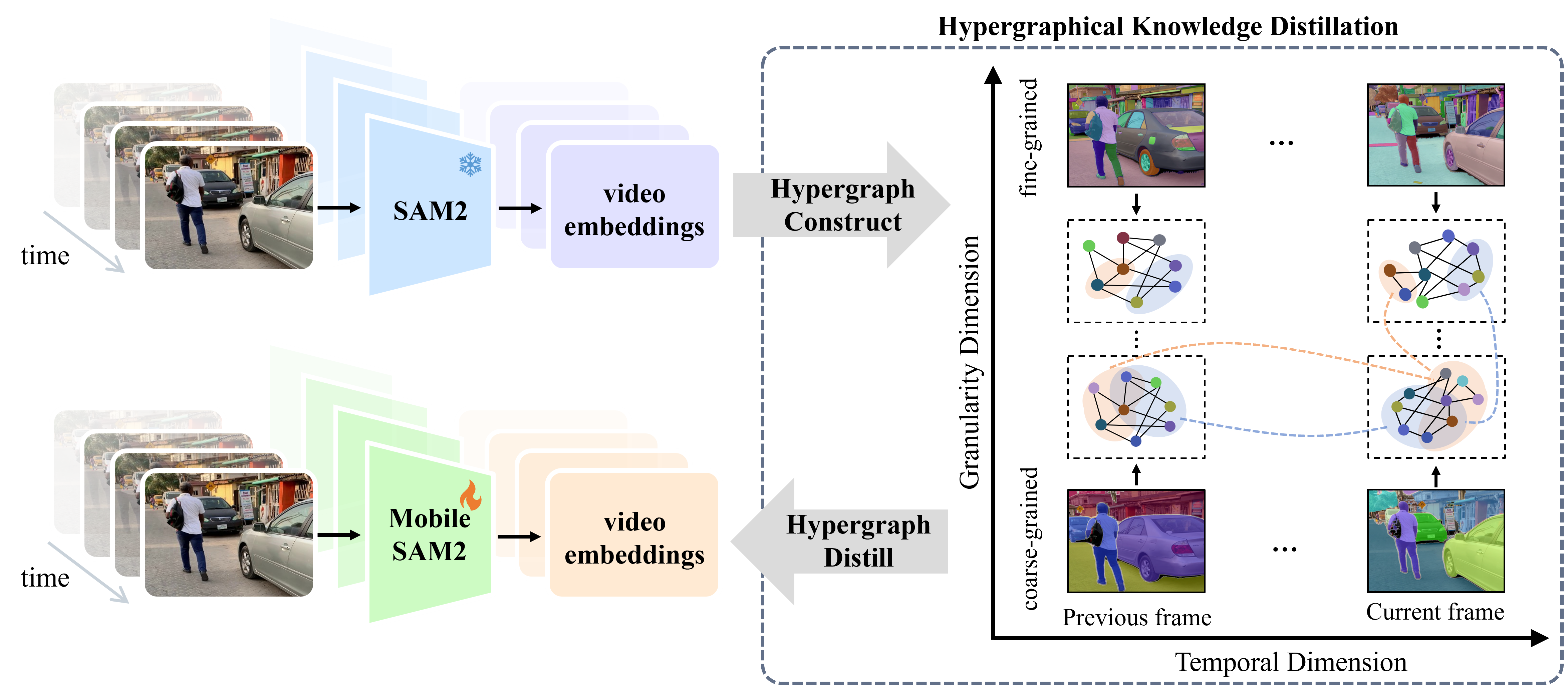}
\vspace{-2.0em}
\caption{
\textbf{Overview of HyperKD.} HyperKD constructs hypergraphs to explicitly model and extract the generalizable temporal knowledge and the comprehensive multi-granularity knowledge from SAM2, which are then distilled into lightweight MobileSAM2 by aligning it with the constructed hypergraphs.
Specifically, Temporal HyperKD considers objects within a single frame as nodes and constructs hypergraphs by linking multiple related nodes across frames via hyperedges, as shown along the temporal dimension. 
Granularity HyperKD treats segmentation entities within a single granularity level as nodes and builds hypergraphs by linking multiple relevant nodes across granularity levels (e.g., objects, parts and subparts) via hyperedges, as shown along the granularity dimension.
In this way, the trained MobileSAM2 captures rich and diverse hypergraphical correlations across multiple frames and granularity levels, ultimately achieving robust and comprehensive video understanding and segmentation.
}
\vspace{-2em}
\label{fig:schematic}

\end{figure*}

\subsection{Preliminaries}\label{sec: Preliminaries}
\textbf{Segment Anything Model 2 (SAM2)}~\cite{ravi2024sam2} is a unified model for video and image segmentation. 
SAM2 is designed to segment anything by utilizing a ``Promptable Segmentation'' scheme, 
where the model generates a segment of spatio-temporal mask (i.e., a `masklet') based on given prompts on any frames of the video, such as points, boxes, or masks. 
SAM2 can segment the object of interest in single image and across video frames when appropriate prompts are provided. 
Additionally, SAM2 support ``interactive segmentation'', which allows users to iteratively refine segments and scale up training data through model-in-the-loop annotation.
This process results in a more powerful SAM2 that can handle diverse segmentation tasks with improved flexibility and accuracy.

SAM2 typically works with two stages.
It first encodes an input frame and a set of prompts into feature embeddings and then predicts the expected segmentation mask conditioned on these features and the memory context from previously observed frames.
Specifically, SAM2 consists of an image encoder $\textbf{Encoder}^{I}$, 
a memory module $\textbf{MemModule}$,
a prompt encoder $\textbf{Encoder}^{P}$, 
and a mask decoder $\textbf{Decoder}$.
Given an input video $x^{I}\in \mathbb{R}^{T\times H\times W\times 3}$, where $T$ is the number of frames, 
and a prompt $x^{P}$ on the first frame, SAM2 first encodes $x^{I}$ and $x^{P}$ into embeddings as follows:

\begin{equation}
         z^{I} = \textbf{Encoder}^{I}(x^{I}), \;
         z^{P} = \textbf{Encoder}^{P}(x^{P}).
\end{equation} 
The memory module then conditions the $i$-th frame embedding $z^{I}(i)$ on the past $N$ frames embeddings $\{z^{I}(j)\}_{j=i-N}^i$ and predictions $\{m(j)\}_{j=i-N}^i$ as follows:
\begin{equation}
\small
z^{I}_m(i) = \textbf{MemModule}\left(z^{I}(i) \mid \{z^{I}(j)\}_{j=i-N}^{i-1}, \{m(j)\}_{j=i-N}^{i-1}\right),
\end{equation} 
where $\{z^{I}(i)\}_{i=t-N}^{t-1}$ and $\{m(i)\}_{i=t-N}^{t-1}$ interact with $z^{I}(t)$, leading to a conditioned frame embedding $z^{I}_m$.

The mask decoder $\textbf{Decoder}$ then decodes the conditioned frame embedding $z^{I}_m(t)$ conditioned on the prompt embedding $z^{P}$:
\begin{equation}
    m(i), c(i) = \textbf{Decoder}\left(z^{I}_m(i)\mid z^{P}\right),
\end{equation}
where $ z^{P} $ interacts with $ z^{I}_m(i) $, producing binary segmentation predictions $ m \in \mathbb{R}^{T \times M \times H \times W} $ that include $M$ valid masks, each corresponding to different levels of segmentation granularity, along with the corresponding confidence scores $ c \in \mathbb{R}^{T \times M} $.

\textbf{Na\"ive Solution with FitNet~\cite{romero2014fitnets}.}
In this paper, we adopt SAM2~\cite{voigtlaender2020siam} as the pretrained teacher model, using Hiera-L~\cite{ryali2023hiera} as image encoder. 
We employ the FitNet approach~\cite{romero2014fitnets} for knowledge distillation, where a teacher model guides a student model by aligning the student’s intermediate feature representations from the image encoder with those from the teacher.
Given an unlabeled video ${x}^I$, the teacher model first generates feature embeddings through its image encoder, which serve as alignment targets for the corresponding layers in the student model’s image encoder. This alignment of intermediate representations enhances knowledge transfer in our distillation setup.
The unsupervised training of the student model on unlabeled data is formulated as:
\begin{equation}\label{fitnetloss}
    \text{Loss} =\|z^I_t - z^I_s\|_1,
\end{equation} 
which aligns the intermediate feature embedding $z^I_t$ and $z^I_s$ from the teacher’s image encoder $\textbf{Encoder}^I_t$ and the student’s image encoder $\textbf{Encoder}^I_s$, respectively.

\subsection{Hypergraphical Knowledge Distillation}
\label{sec: Hypergraphical Knowledge Distillation}

We observe that SAM2’s success is driven by its captured two key types of knowledge from large-scale, multi-granular video data: 
temporal knowledge for robust segmentation across frames and multi-granularity knowledge for detailed understanding of visual concepts at multiple levels. 
Motivated by this, we propose Hypergraphical Knowledge Distillation (HyperKD), which uses hypergraphs in knowledge distillation to effectively capture and encapsulate above two types of knowledge into a lightweight model. 
Since SAM2’s temporal and multi-granularity knowledge are implicitly encoded in its parameters, HyperKD explicitly models them through hypergraphs, 
which the student model learns to emulate. 
We instantiate HyperKD with two methods: 
Temporal HyperKD that links related nodes across frames to capture temporal consistency, and Granularity HyperKD that connects nodes across segmentation levels to convey multi-level understanding.

\subsubsection{Temporal HyperKD}
Temporal HyperKD works in a two-step manner.
The first step is temporal hypergraph extraction with node and hyperedge construction that aims to uncover the implicitly encoded  generalizable temporal knowledge in SAM2. 
The second step is temporal hypergraph encapsulation, which encapsulates the extracted temporal hypergraph into the student model, enabling the student model to generate more temporally consistent segmentation.

\textbf{Temporal Hypergraph Extraction.}
With the generalizable temporal knowledge contained in video embedding $z^{I}_t$ from teacher model, 
we formulate the proposed Patch-wise Temporal Hypergraph and Instance-wise Temporal Hypergraph as $\mathcal{G}^{Pa}_t = (\mathcal{V}^{Pa}_t, \mathcal{E}^{Pa}_t)$ and $\mathcal{G}^{Ins}_t = (\mathcal{V}^{Ins}_t, \mathcal{E}^{Ins}_t)$
, which are capable of capturing temporal semantic relationships and associations across frames.
For $\mathcal{G}^{Pa}_t$, we deconstruct the video embedding $z^{I}_t$ as grid-based patches to constitute vertex set $\mathcal{V}^{Pa}_t$. 
For $\mathcal{G}^{Ins}_t$, we constitute vertex set $\mathcal{V}^{Ins}_t$ by pooling the video embedding $z^{I}_t$ according to high confidence segmentation entries of each frame in teacher model prediction $m_t$.

Without loss of generality, to establish hyperedges that model relationships among vertices in $\mathcal{V}$, 
we apply an $ \epsilon $-ball distance threshold around each vertex. 
An $ \epsilon $-ball forms a hyperedge that includes all vertices within a certain distance $ \epsilon $ from a central vertex. 
The set of hyperedges $ \mathcal{E}^{Pa}_t$ or $\mathcal{E}^{Ins}_t $ are therefore constructed as follows:
\begin{equation}
\mathcal{E} = \{ \text{ball}(v, \epsilon) \mid v \in \mathcal{V} \},
\label{eq:HyperedgeExtraction}
\end{equation}
where $\text{ball}(v, \epsilon) = \{ u \mid \text{dist}(x^I_u - x^I_v) < \epsilon, \, u \in \mathcal{V} \}$
represents the neighboring vertex set of a given vertex $ v $, $\mathcal{V}$ is $\mathcal{V}^{Pa}_t$ or $\mathcal{V}^{Ins}_t$, and $ \text{dist}(x^I_u - x^I_v) $ is the distance function used to determine proximity. 
In practical, the hypergraph $ \mathcal{G} $ is represented by its incidence matrix $ H $, 
where $ H(u,v) = \text{dist}(x^I_u - x^I_v) = 1-\langle x^I_u, x^I_v\rangle/(\|x^I_u\|\|x^I_v\|)$ if vertex $ u $ is part of the hyperedge centered at vertex $ v $, 
and $ H(u,v) = 0 $ otherwise.

\textbf{Temporal Hypergraph Encapsulation}
encapsulates both patch-level and instance-level generalizable temporal knowledge captured by the extracted hypergraph $\mathcal{G}^{Pa}_t$ and $\mathcal{G}^{Ins}_t$ into the student model, thereby preserving SAM2’s generalizable temporal knowledge within the student model.
Specifically, similar to the construction of $ \mathcal{G}^{Pa}_t $ and $ \mathcal{G}^{Ins}_t $, we construct $ \mathcal{G}^{Pa}_s $ and $ \mathcal{G}^{Ins}_s $ for video embeddings $ z^{I}_s $ from student model.
Then, we encapsulate the extracted temporal hypergraph $\mathcal{G}^{Pa}_t $ and $ \mathcal{G}^{Ins}_t $ into student model by minimizing:
\begin{equation}
    \mathcal{L}_\text{THKD} = \|H^{Pa}_t-H^{Pa}_s\|_\text{F} + \|H^{Ins}_t-H^{Ins}_s\|_\text{F},
\end{equation}
where $H^{Pa}_t$, $H^{Ins}_t$, $H^{Pa}_s$, and $H^{Ins}_s$ refer to the incidence matrices of $\mathcal{G}^{Pa}_t$, $\mathcal{G}^{Ins}_t$, $\mathcal{G}^{Pa}_s$, and $\mathcal{G}^{Ins}_s$, respectively.

In this way, Temporal HyperKD ensures that both patch-level and instance-level temporal hypergraphs capture neighborhood relationships within the feature space, effectively distilling SAM2’s generalizable temporal knowledge into the student model, enabling the student model to maintain temporal consistency and achieve robust segmentation across video frames.



\subsubsection{Granularity HyperKD}

As Temporal HyperKD distills only the generalizable temporal knowledge, 
we further design Granularity HyperKD to extract a granularity hypergraph and encapsulate it into the student model to improve segmentation precision across different levels of detail. 
Specifically, the granularity hypergraph captures hierarchical relationships between multi-granularity level segmentation entities, complementing the temporal hypergraph by providing orthogonal and multi-granularity knowledge.

\textbf{Granularity Hypergraph Extraction.}
Given the $i$-th frame embedding $z^{I}_t(i)\in \mathbb{R}^{D\times h\times w}$ and its predicated binary segmentation masks $m(i)\in \mathbb{R}^{M\times H\times W}$ with $M$ levels of segmentation granularity from teacher model, 
we instantiate granularity hypergraph as $\mathcal{G}^{Gran}_t = (\mathcal{V}^{Gran}_t, \mathcal{E}^{Gran}_t)$, in which each node ${v}\in \mathcal{V}^{Gran}_t$ is initialized by pooling the frame embedding according to predicated binary segmentation masks as follows:
\begin{equation}
\footnotesize
\mathcal{V}^{Gran}_t=\left\{\textbf{MaskedAveragePooling}\left(z^{I}_t(i,j), m(i,j)\right)\mid j=1,\dots, M \right\},
\end{equation}
and the set of hyperedges $\mathcal{E}^{Gran}_t$ can be constructed with Eq.~\ref{eq:HyperedgeExtraction}.

\textbf{Granularity Hypergraph Encapsulation}
encapsulates the orthogonal and multi-granularity knowledge in the extracted granularity hypergraph into the student model. This process complements the temporal hypergraph and enhances segmentation precision by enabling the model to segment objects at various levels of detail, such as objects, parts, and subparts.
Specifically, similar to the construction of $ \mathcal{G}^{Gran}_t $, we construct $ \mathcal{G}^{Gran}_s $ for the video embeddings $ z^{I}_s $ from the student model, using the predictions $ m(i) $ from the teacher model.
We then encapsulate the extracted granularity hypergraph into the student model, following a similar approach as in Temporal Hypergraph Encapsulation, by minimizing the following loss function:
\begin{equation}
    \mathcal{L}_\text{GHKD} = \|H^{Gran}_t-H^{Gran}_s\|_\text{F},
\end{equation}
where $ H^{Gran}_t $ and $ H^{Gran}_s $ are the incidence matrices of $ \mathcal{G}^{Gran}_t $ and $ \mathcal{G}^{Gran}_s $, respectively.

\subsection{Overall Objective}
In summary, the overall training loss of the proposed HyperKD can be formulated as:
\begin{equation}
    \text{Loss} =\|z^I_t - z^I_s\|_1 + \alpha\cdot\mathcal{L}_{THKD} + \beta\cdot\mathcal{L}_{GHKD},
\end{equation}
where $\alpha$ and $\beta$ are weighting coefficients. 

\subsection{Model Architectures}\label{sec: Model Architectures}

We introduce MobileSAM2, a lightweight version of SAM2 that strikes a balance between model size and performance. 
While SAM2's memory module, prompt encoder, and mask decoder are lightweight (under 12M parameters), its image encoder, based on Hiera-L~\cite{ryali2023hiera}, has over 212M parameters, making it too heavy for resource-constrained devices. 
Thus, the key instantiate MobileSAM2 is reducing the size of the image encoder while preserving SAM2's robust and comprehensive video segmentation capabilities.

Motivated by this observation, we present a new family of Hierarchical Vision Transformer~\cite{ryali2023hiera} for SAM2
by 
scaling down a large model seed with a progressive model contraction approach~\cite{feichtenhofer2020x3d}.
Specifically, we begin with a manually designed Hiera backbone that serves as image encoder, which maintaining the core principles of the original Hiera design.
Then, we establish a parameterized search space of contraction factors that includes critical architectural components such as embedding dimensions, layer counts, attention head configurations, and expansion ratios. 
The progressive model contraction process iteratively identifies promising architectures, ultimately resulting in a smaller Hiera variant that retains competitive performance while minimizing computational complexity and memory usage, making it suitable for efficient deployment.
Hiera begins with a patch embedding layer that processes input images into tokens, followed by several stages where each stage progressively increases the channel dimensions while reducing spatial resolution through downsampling. 
The architecture emphasizes efficient parameter usage by employing lightweight MultiScaleBlock layers in early stages, transitioning to windowed self-attention in later stages. 
We consider the following contraction factors to form a model:
\begin{itemize}
    \item $\Gamma_\text{Emb}$: {Embedding dimension of each stage.} Decreasing it results in a thinner network with fewer heads in multi-head self-attention.   
    \item $\Gamma_\text{Blk}$: {The number of blocks in each stage.} The depth of the model is decreased by reducing these values. 
    \item $\Gamma_\text{WinSiz}$: {Window size in the each stage.} 
    Smaller window size lead to fewer operations and model parameters during self-attention. 

    \item $\Gamma_\text{HExp}$: {Expansion ratio of attention heads in multi-head attention at each stage.} Reducing the dimensionality of each attention head directly decreases the computation burden of self-attention, leading to lower computation cost.
\end{itemize}

We scale down the above factors by the progressive model contraction approach and search and identify a new family of lightweight SAM2, named MobileSAM2, as shown in Table~\ref{tab:ModelContractionResult}. 

\begin{table}[t]
\small
  \centering
    \centering	
     \caption{
     Architectures of our searched lightweight MobileSAM2\protect\footnotemark. 
     }
    \vspace{-1em}
    \begin{tabular}	{c | cccc|c } 
    \midrule
    Contraction Factors&$\Gamma_\text{Emb}$ & $\Gamma_\text{Blk}$ & $\Gamma_\text{WinSiz}$ &$\Gamma_\text{HExp}$ &Params. (M) \\
    \midrule
    MobileSAM2-5M& [48, 96, 192, 384] &[1, 2, 5, 2]& [8, 4, 14, 7]& 2& 5.84	   \\
    MobileSAM2-10M& [64, 128, 256, 512] &[1, 2, 5, 2] & [8, 4, 14, 7]& 2&10.37    \\
    MobileSAM2-23M& [96, 192, 384, 768] &[1, 3, 9, 1] & [8, 8, 14, 7] & 2 & 23.74 \\
    \bottomrule
    \end{tabular}
    \vspace{-1.5em}
  \label{tab:ModelContractionResult}
\end{table}
\footnotetext{Please refer to appendix for more details of MobileSAM2 architecture.}

\begin{table*}[t]
\centering
\caption{
\textbf{VOS comparison.}
With the proposed HyperKD and the searched model architectures, our MobileSAM2 works effectively with limited parameters on video segmentation. 
Note SAM2 Large~\cite{ravi2024sam2} serves as the teacher model for all knowledge distill methods.
For fair comparsion, 
all methods process video sequences.
}
\vspace{-1em}
\resizebox{\linewidth}{!}{
\begin{tabular}{l c c ccccc }
\toprule
\multirow{2}*{{Model}}  &\multirow{2}*{{Distillation Method}} &\multirow{2}*{Params. (M)}&  \multicolumn{5}{c}{J\&F} \\
\cmidrule(r){4-8}
&&& MOSE val & DAVIS 2017 val & LVOS val & SA-V val & SA-V test \\
\midrule
\multicolumn{8}{c}{100\% Training Data with 256 A100 GPUs: 113.8K Videos (SA-V manual and internal data~\cite{ravi2024sam2})}  \\
\midrule
SAM2 Base+~\cite{ravi2024sam2} & - &68.7&72.8&88.8&75.8&72.2&74.7 \\
SAM2 Large~\cite{ravi2024sam2} & - &212.1 &74.6&89.2&81.7&74.5&76.0 \\
\midrule
\multicolumn{8}{c}{$\sim$ 10\% Training Data with 1 A100 GPU: 11K Videos from SA-V manual data~\cite{ravi2024sam2})}\\
\midrule

\multirow{2}*{SAM2-TinyViT-5M}
&No Distill &5.4&26.0&30.4&30.8&28.4&27.7\\
&FitNet~\cite{romero2014fitnets} &5.4&33.8&38.7&37.7&34.3&35.1\\
\cmidrule(r){2-8}
\multirow{6}*{MobileSAM2-5M}
&No Distill &5.8&29.7&36.6&33.8&32.7&33.0\\
&FitNet~\cite{romero2014fitnets} &5.8&36.6&41.5&39.7&38.2&39.3\\ 
&CIRKD~\cite{yang2022cross} &5.8&36.8&43.3&41.2&38.3&38.6\\ 
&CAT-KD~\cite{guo2023class} &5.8&35.0&41.9&40.3&38.3&39.8\\ 
&FAKD~\cite{yuan2024fakd} &5.8&37.7&42.5&41.0&39.6&39.9\\ 
\rowcolor{gray!16}&HyperKD (Ours) &5.8&44.6&51.0&48.8&49.2&49.4\\

\midrule

\multirow{2}*{SAM2-TinyViT-11M}
&No Distill &11.2&27.9&29.6&32.5&28.9&30.8\\
&FitNet~\cite{romero2014fitnets} &11.2&36.4&40.0&39.8&37.8&39.0\\
\cmidrule(r){2-8}
\multirow{6}*{MobileSAM2-10M}
&No Distill &10.4&30.9&35.7&34.4&33.7&34.2\\
&FitNet~\cite{romero2014fitnets} &10.4&38.9&44.3&42.2&41.6&41.4\\ 
&CIRKD~\cite{yang2022cross} &10.4&40.8&45.8&43.7&42.4&43.2\\ 
&CAT-KD~\cite{guo2023class} &10.4&39.9&44.3&42.9&41.3&41.8\\ 
&FAKD~\cite{yuan2024fakd} &10.4&41.3&46.7&44.0&42.4&43.3\\ 
\rowcolor{gray!16}&HyperKD (Ours) &10.4&48.7&54.2&51.8&49.9&50.0\\
\midrule

\multirow{2}*{SAM2-TinyViT-21M}
&No Distill &21.2&30.3&35.2&34.5&32.0&34.7\\
&FitNet~\cite{romero2014fitnets} &21.2&44.9&49.0&47.0&45.8&45.6\\
\cmidrule(r){2-8}
\multirow{6}*{MobileSAM2-23M}
&No Distill &23.7&39.5&44.7&44.8&42.4&44.3\\
&FitNet~\cite{romero2014fitnets} &23.7&57.4&61.1&60.8&59.4&59.9\\ 
&CIRKD~\cite{yang2022cross} &23.7&60.4&62.1&61.5&60.7&60.5 \\ 
&CAT-KD~\cite{guo2023class} &23.7&60.2&62.1&61.1&59.7&59.9\\ 
&FAKD~\cite{yuan2024fakd} &23.7&60.6&65.0&63.4&62.0&62.7\\ 
\rowcolor{gray!16}&HyperKD (Ours) &23.7&65.8&72.1&69.0&66.4&67.8\\

\bottomrule
\end{tabular}
}

\label{tab:result}
\vspace{-1em}
\end{table*}

\section{Experiments}

Table~\ref{tab:result} show the benchmarking of our methods with state-of-the-art knowledge distillation methods, including FitNets~\cite{romero2014fitnets}, CIRKD~\cite{yang2022cross}, and  
FAKD~\cite{yuan2024fakd}, over 5 widely used video segmentation datasets including 2 general video object segmentation (VOS) dataset (i.e., MOSE~\cite{ding2023mose}, DAVIS 2017~\cite{voigtlaender2020siam}), 1 long-term VOS dataset (LVOS~\cite{hong2023lvos}), and 2 segment anything in videos dataset (SA-V)~\cite{ravi2024sam2} (i.e., dataset SA-V-val~\cite{ravi2024sam2} and SA-V-test~\cite{ravi2024sam2}).

\subsection{Implementation Details}\label{sec:Implementation Details}
We use our designed MobileSAM2 as the lightweight student model and SAM2-L~\cite{voigtlaender2020siam} (with Hiera-L~\cite{ryali2023hiera} as the image encoder) as the pretrained teacher model.
Since SAM2's memory module (i.e., memory attention, memory encoder, and memory bank), prompt encoder and mask decoder are relatively lightweight (under 12M parameters),
we optimize MobileSAM2 by training its image encoder while keeping its remaining modules (i.e., memory module, prompt encoder and mask decoder copied from the pretrained SAM2-L teacher model) frozen.
We use only $\sim10\%$ SA-V data~\cite{ravi2024sam2} (i.e., 11K Videos) for efficient distillation training. 
Following SAM2~\cite{ravi2024sam2}, we employ the AdamW~\cite{loshchilov2017decoupled} optimizer with an initial learning rate of $5 \times 10^{-4}$. 
All models are trained for 5 epochs on an A100 (80GB) GPU with a batch size of 5, using 8-frame sequences per sample. 
The training time is 65 hours. 
For each iteration, we use a $4 \times 4$ grid of spatial prompts for mask prediction by the teacher model.
We set weighting coefficients $\alpha=1$ and $\beta=1$ to balance temporal consistency and hierarchical knowledge transfer.

\subsection{Results}\label{sec:exp_result}

In line with SAM2~\cite{ravi2024sam2}, our proposed MobileSAM2 is designed for general, interactive, promptable video segmentation tasks, while we also address the semi-supervised video object segmentation (VOS) setting, where the prompt is a ground-truth mask on the first frame, as this is a common protocol in the field. 
We compare MobileSAM2 with existing state-of-the-art knowledge distillation methods as well as SAM2 equipped with the state-of-the-art lightweight backbone (Tiny-ViT~\cite{wu2022tinyvit}) in Table~\ref{tab:result}, reporting accuracy based on standard protocols.
We evaluate three versions of MobileSAM2 with varying model sizes (5M, 10M, and 23M), each offering different size-accuracy trade-offs. 


\textbf{Results on general VOS dataset.}
Table~\ref{tab:result} presents the video object segmentation results on two general VOS datasets: MOSE and DAVIS 2017. 
MobileSAM2 achieves substantial performance improvements over the baseline across these datasets. 
Specifically, MobileSAM2 outperforms state-of-the-art methods by a large margin in J\&F score, 
highlighting the effectiveness of our proposed HyperKD in exploring efficient lightweight model architectures as well as distilling knowledge from the pretrained SAM2 model.
The superior performance of MobileSAM2 is largely attributed to HyperKD’s ability to capture generalizable temporal knowledge and comprehensive multi-granularity knowledge, enabling it to focus on cross-temporal and multi-granularity spatial cues of objects. 
This not only benefits the architecture search for lightweight MobileSAM2 but also enhances the distillation process from SAM2 to lightweight MobileSAM2 .
All methods show performance gains. 
However, MobileSAM2 achieves the most significant improvements, 
demonstrating its capability to align closely with the SAM2’s representation by effectively distilling both temporal and multi-granularity knowledge from SAM2.

\begin{table}[t]
\centering
\caption{
Ablation studies of HyperKD with Temporal HyperKD and Granular HyperKD. 
The experiments are conducted on video object segmentation (VOS) over LVOS val.}
\vspace{-1em}
\resizebox{\linewidth}{!}{
\begin{tabular}{l|c| c|c | c}
\toprule
&\multicolumn{2}{c|}{Temporal HyperKD} &\multirow{2}*{Granular HyperKD} &\multirow{2}*{{J}\&{F}}\\
\cmidrule(r){2-3}
& Patch-wise&Instance-wise& \\
\midrule
MobileSAM2-23M (Baseline/no distill) &-&-&-&44.8\\
\midrule
& \checkmark&&&62.9\\
&&\checkmark &&64.4\\
& \checkmark&\checkmark &&64.4\\
&&&\checkmark &63.1\\
\rowcolor{gray!16} MobileSAM2-23M & \checkmark&\checkmark &\checkmark &69.0\\
\bottomrule
\end{tabular}
}
\label{tab:ablation_study}
\vspace{-1em}
\end{table}

\textbf{Results on long-term VOS dataset.}
Table~\ref{tab:result} also reports the experiments on long term VOS dataset, LVOS.
MobileSAM2 surpasses all competing methods by a significant margin, demonstrating HyperKD’s effectiveness and robustness in capturing SAM2’s generalizable temporal and multi-granularity knowledge for long-term video segmentation. 
The substantial performance gains achieved by MobileSAM2 on LVOS highlight the efficiency of its lightweight architecture, explored by HyperKD’s guidance, as well as the effectiveness of distilling knowledge from SAM2 by HyperKD.

\textbf{Results on segment anything in videos dataset.}
We evaluate the effectiveness of our MobileSAM2 on segment anything in videos dataset, i.e., SA-V val and SA-V test, which measure performance for open-world segments of “any” object class~\cite{ravi2024sam2}.
Table~\ref{tab:result} reports the VOS result, showcasing significant improvements over the {baseline} and outperforming state-of-the-arts thereby highlighting the superiority of MobileSAM2.

\subsection{Ablation Study}\label{sec:exp_ablation}
In Table~\ref{tab:ablation_study},  
we conduct ablation studies to assess the individual contributions of our proposed MobileSAM2 on the LVOS dataset. 
The baseline model, without distillation from SAM2, does not perform well due to its limited ability to capture high-order temporal and multi-granularity semantic relationships of visual concepts. 
By contrast, including Patch-wise Temporal HyperKD significantly improves the baseline, indicating that Patch-wise Temporal HyperKD effectively captures patch-wise temporal dependencies across frames. 
Additionally, applying Instance-wise Temporal HyperKD brings further improvements, showing that it enhances instance-level temporal knowledge across frames. 
Moreover, introducing Granularity HyperKD further boosts performance, demonstrating that it provides complementary multi-granularity knowledge that effectively benifits robust and comprehensive video segmentation capabilities.

\subsection{Discussion}\label{sec:exp_discussion}
\textbf{Parameter Study.}
In the construction of hyper graph, the predefined $ \epsilon $-ball distance threshold is used to establish hyperedges that model relationships among vertices in hypergraph.
We studied $\epsilon$ by changing it from $0.5$ to $1.5$ with a step of $0.25$. 
Table~\ref{tab:para_study_epsilon} reports the experiments over LVOS dataset. 
It is also observed that the performance of MobileSAM2 is relatively stable across from 0.75 to 1.25, with only minor variations,
while there is a performance decline at the thresholds of 0.5 and 1.5.
A higher threshold creates a more connected hypergraph, increasing the risk of over-smoothing, while a lower threshold may result in an under-connected hypergraph that fails to capture high-order relationships. To balance these factors, our HyperKD uses a distance threshold of 1.0 for hypergraph construction, chosen empirically to maintain connectivity without excessive smoothing.

\begin{table}[t]
\centering
\caption{Parameter analysis on hypergraph construction threshold $\epsilon$ in HyperKD on LVOS.}
\vspace{-1em}
\footnotesize
\begin{tabular}{c|ccccc}
\toprule
Model Architecture & \multicolumn{5}{c}{MobileSAM2-23M} \\
\midrule
$\epsilon$     &0.5      &0.75      &\cellcolor{gray!16}{1.0}      &1.25      &1.5\\
\midrule
J\&F    &65.2 &68.4    &\cellcolor{gray!16}69.0   &68.1  &62.0 \\
\bottomrule
\end{tabular}
\label{tab:para_study_epsilon}
\vspace{-1em}
\end{table}

\textbf{HyperKD v.s. Graph-based Knowledge Distillation.}
We compare our HyperKD with the prior graph-based knowledge distillation method, Context Matters~\cite{yang2023context} and IntRA-KD~\cite{hou2020inter}.  
As shown in Table~\ref{tab:HyperKD v.s. Graph-based Knowledge Distillation}, HyperKD outperforms Context Matters and IntRA-KD, 
primarily because the graphs of Context Matters and IntRA-KD cannot effectively capture high-order semantic relationships among visual concepts, 
which our hypergraph-based approach successfully achieves.

\begin{table}[t]
\centering
\caption{
Comparison of our HyperKD with traditional two-vertices graph-based knowledge distillation over LVOS. 
}
\vspace{-1em}
\scriptsize
\resizebox{0.98\linewidth}{!}{
\begin{tabular}	{c| c|ccc } 
\toprule
Model Architecture & \multicolumn{4}{c}{MobileSAM2-23M} \\
\midrule

Distillation Method & No Distill & Context Matters~\cite{yang2023context} &IntRA-KD~\cite{hou2020inter} & HyperKD (Ours)\\
\midrule
{J\&F} &44.8&60.7&60.9 &\cellcolor{gray!16}{69.0} \\
\bottomrule
\end{tabular}
}
\label{tab:HyperKD v.s. Graph-based Knowledge Distillation}
\end{table}

\begin{table}[t]
\centering
\caption{
Efficiency comparison on mobile GPU.
}
\vspace{-1em}
\footnotesize
\begin{tabular}	{c| cc c} 
\toprule
Model & Params. (M) & Inference FPS &LVOS \\
\midrule
SAM2 Base+ &68.7 & {Out of Memory} &-\\
SAM2 Large &212.1 & {Out of Memory} &-\\
\midrule
SAM2-TinyViT-5M &5.4 &13.1	 &37.7\\
\rowcolor{gray!16} MobileSAM2-5M &5.8	& 13.3   &48.8\\
SAM2-TinyViT-11M &11.2	& 11.3 & 39.8\\
\rowcolor{gray!16} MobileSAM2-10M &10.4	& 10.8 &51.8\\
SAM2-TinyViT-21M &21.2	& 8.0 & 47.0\\
\rowcolor{gray!16} MobileSAM2-23M &23.7	& 8.2 &69.0\\
\bottomrule
\end{tabular}
\label{tab:computational_overhead}
\vspace{-1.0em}
\end{table}

\textbf{Distance Metrics for Constructing Hypergraph.}
We explore different feature distance metrics for hypergraph construction, 
conducting experiments with the following metrics: 1) Cosine Similarity~\cite{deza2009encyclopedia}, 2) Euclidean Distance~\cite{deza2009encyclopedia}, and 3) Manhattan Distance~\cite{deza2009encyclopedia}. 
The results in \cref{tab:Distance_Metrics_Study} show that HyperKD performs effectively and consistently across all metrics. 
Among them, cosine similarity yields the best results, 
largely because it aligns with the attention mechanism, making it a natural choice for the distillation of generalizable temporal knowledge and comprehensive multi-granularity knowledge in SAM2.

\begin{table}[t]
\centering
\caption{
Analysis on hypergraph construction in HyperKD with different distance metrics on LVOS.
}
\vspace{-1em}
\resizebox{\linewidth}{!}{
\begin{tabular}{l|c|c|c}
\toprule
Model Architecture & \multicolumn{3}{c}{MobileSAM2-23M} \\
\midrule
Distance Metrics     &{Cosine Similarity (Ours)}      &Euclidean Distance      &Manhattan Distances\\
\midrule
J\&F    &\cellcolor{gray!16}69.0   &67.2  &65.7\\
\bottomrule
\end{tabular}
}
\label{tab:Distance_Metrics_Study}
\vspace{-2em}
\end{table}

\textbf{Efficiency Analysis.}
We evaluate model efficiency in inference speed for all models on a single RTX 4060 (8GB) mobile GPU using PyTorch 2.3.1 and CUDA 12.1 with automatic mixed precision (bfloat16). 
We compiled the image encoder with torch.compile for all models. 
FPS measurements for video object segmentation are taken with a batch size of 1, following the common protocol.
Table~\ref{tab:computational_overhead} reports the results on LVOS dataset, demonstrating that MobileSAM2 achieves a significant reduction in computational overhead while maintaining competitive performance compared to the original SAM2, making it more mobile-friendly for resource-constrained devices.


\begin{table}[t]
    \centering
    \caption{Applicability of HyperKD for TrackAnything (TAM)~\cite{yang2023track}. We report J\&F scores on DAVIS-2017.
    }
    \label{tab:trackanything_kd}
    \vspace{-1em}
    \scriptsize
    \begin{tabular}{lccc}
        \toprule
        Model &Distillation Method &Params. (M) & DAVIS-2017 J\&F\\
        \midrule
        \multicolumn{4}{c}{\textit{$100\%$ Training Data}} \\
        \midrule
        TrackAnything (TAM)~\cite{yang2023track} &-&636& 73.1 \\
        \midrule
        \multicolumn{4}{c}{\textit{$\sim 10 \%$ Training Data}} \\
        \midrule
        \multirow{2}*{TAM-TinyViT-21M} & No Distill &21.2& 58.9 \\
        &\cellcolor{gray!16}HyperKD (Ours) &\cellcolor{gray!16}21.2 & \cellcolor{gray!16}\textbf{63.1}\\
        \cmidrule(r){2-4}
        \multirow{2}*{MobileTAM-23M} & No Distill &23.7 & 60.8\\
        &\cellcolor{gray!16}HyperKD (Ours) &\cellcolor{gray!16}23.7 & \cellcolor{gray!16}\textbf{66.5} \\
        \bottomrule
    \end{tabular}
    \vspace{-1em}
\end{table}

\textbf{Applicability of HyperKD to Other Foundation Models.} 
We conduct experiments on TrackAnything (TAM)~\cite{yang2023track} as shown in Table~\ref{tab:trackanything_kd}. 
It demonstrates HyperKD's applicability beyond SAM2: applied to the structurally independent TrackAnything (TAM)~\cite{yang2023track} under the same limited-data setting ($\sim$10\%), HyperKD improves TAM-TinyViT-21M from 58.9 to 63.1 J\&F and MobileTAM-23M from 60.8 to 66.5 J\&F on DAVIS-2017. 
More critically, it disentangles the distillation objective from the architecture design: since TAM shares no architectural lineage with SAM2 or MobileSAM2, the consistent gains confirm that the performance improvements stem from the HyperKD objective itself, not from co-adaptation with a specific backbone. Together, these findings establish HyperKD as a generalizable distillation framework, and validate that the gains on MobileSAM2 reflect genuine knowledge transfer rather than artifacts of the searched architecture.

\begin{table}[t]
    \centering
    \caption{Application of MobileSAM2 on Embodied AI.
    }
    \label{tab:vls}
    \vspace{-1em}
    \scriptsize
    \resizebox{\linewidth}{!}{
    \begin{tabular}{lccccccc}
        \toprule
        Method &\multicolumn{2}{c}{Libero-Object}		&\multicolumn{2}{c}{Libero-Goal}		&\multicolumn{2}{c}{Libero-Spatial} &Avg.\\
        &Pos (Avg.)	&Task (Avg.)	&Pos (Avg.)	&Task (Avg.)	&Pos (Avg.)	&Task (Avg.)	\\
        \midrule
        OpenVLA~\cite{kim2024openvla} &0 &0 &0 &0 &0 &0 &0\\
        $\pi$0~\cite{black2024pi_0} &0 &0 &0 &0 &0 &0 &0\\
        $\pi$0.5~\cite{black2025pi_}&17 &1 &38 &0 &20 &1 &12.8\\
        SAM3\cite{carion2025sam3segmentconcepts}-CaP-Agent0~\cite{fu2026cap} &27&31&29&16&13&23&23.2\\
        \midrule
        SAM2-TinyViT-21M - CaP-Agent0 &21&27&23&13&11&21&19.3\\
        \rowcolor{gray!16}MobileSAM2-23M - CaP-Agent0 &24&30&28&14&12&23&21.8\\
        \midrule
        \bottomrule
    \end{tabular}
    }
    \vspace{-1.5em}
\end{table}

\textbf{Application of MobileSAM2 on Embodied AI.} 
With the advent of reasoning models and agents~\cite{fu2026cap,Zhang_2025_ICCV,yao2026mulberry}, to verify the application of MobileSAM2 on embodied AI systems, we integrate MobileSAM2 into CaP-Agent0~\cite{fu2026cap} as the core perception model and evaluate manipulation performance on LIBERO-PRO~\cite{2025liberpro}.
As shown in Table~\ref{tab:vls}, the full SAM3-powered CaP-Agent0 achieves 23.2\% average success rate, outperforming training-based VLA baselines like \(\pi_{0.5}\)~\cite{black2025pi_} (12.8\%). 
At a comparable lightweight parameter scale, SAM2-TinyViT-21M built on the state-of-the-art Tiny-ViT~\cite{wu2022tinyvit} backbone drops to 19.3\%. 
In comparison, our MobileSAM2-23M reaches 21.8\% average success rate, retaining competitive performance and outperforming SAM2-TinyViT-21M on spatial reasoning tasks and embodied AI tasks.
These results demonstrate that MobileSAM2 is an efficient perceptual backbone for resource-constrained robot platforms, 
enabling practical on-board deployment with minimal performance sacrifice.

\section{Conclusion}
This paper presents a new family of tiny and efficient video foundation models, MobileSAM2, distilled from large SAM2, using the proposed HyperKD.
HyperKD consists of Temporal HyperKD and Granularity HyperKD that construct hypergraphs to explicitly model and extract the generalizable temporal knowledge and the comprehensive multi-granularity knowledge from SAM2 respectively, facilitating the knowledge distillation from SAM2 to MobileSAM2 as well as the model architecture searching of MobileSAM2.
Extensive experiments on multiple video segmentation datasets demonstrate that MobileSAM2 consistently outperforms state-of-the-art techniques by clear margins. 

\section*{Acknowledgment}
[Re Prof Tao] This project is supported by the National Research Foundation, Singapore, under its NRF Professorship Award No. NRF-P2024-001.
This work is also supported by PolyU Internal Fund.

\bibliographystyle{splncs04}
\bibliography{main}

\clearpage

\section{Appendix}
\subsection{Model Architectures}
Our proposed MobileSAM2 architecture is detailed in \Cref{tab:MobileSAM2Architecture}. This architecture follows a hierarchical structure, starting with a patch embedding layer and progressing through four stages. To create a compact family of MobileSAM2 models, we employ contraction factors $\left\{\Gamma_\text{Emb}, \Gamma_\text{Blk}, \Gamma_\text{WinSiz}, \Gamma_\text{HExp}\right\}$, which enable us to scale down the model size.
We begin with a base model containing 27M parameters, then generate a set of candidate models by adjusting these contraction factors. From these candidates, we select models that meet specific constraints on parameter count and throughput. We evaluate these selected models on approximately 2\% of the SA-V manual dataset~\cite{ravi2024sam2} for training and validate them on the LVOS~\cite{hong2023lvos} validation set.
The models with the highest validation accuracy are further refined in subsequent steps until the target performance is achieved.

\begin{table*}[h]
\small
  \centering
    \caption{
     Architectures of our searched lightweight MobileSAM2\protect\footnotemark. 
     }
    \resizebox{\linewidth}{!}{
    \centering	
    \begin{tabular}	{c| ccccccc} 
    \midrule
    &Stage&Block & Configuration   \\
    \midrule
    \multirow{15}{*}{MobileSAM2}
    &\makecell{Patch\\ Embedding} & Conv & Embed Dim $\Gamma_\text{Emb\_1}$ \\
    \cmidrule{2-4}
    &Stage 1 & Multi-Scale Block~\cite{wu2022tinyvit} &$\begin{bmatrix}
                                    \text{Inupt Dim}\;\Gamma_\text{Emb\_1},\;
                                    \text{Outupt Dim}\;\Gamma_\text{Emb\_2} \\
                                    \text{Window Size}\;\Gamma_\text{WinSiz\_1},\;
                                    \text{Attention Heads}\;1
                                \end{bmatrix}\times 1$\\
    \cmidrule{2-4}
    &Stage 2 & Multi-Scale Block~\cite{wu2022tinyvit} &\makecell{$\begin{bmatrix}
                                    \text{Inupt Dim}\;\Gamma_\text{Emb\_2},\;
                                    \text{Outupt Dim}\;\Gamma_\text{Emb\_2} \\
                                    \text{Window Size}\;\Gamma_\text{WinSiz\_2},\;
                                    \text{Attention Heads}\;\Gamma_\text{HExp}^{2}
                                \end{bmatrix}\times 1$ \\
                                $\begin{bmatrix}
                                    \text{Inupt Dim}\;\Gamma_\text{Emb\_2},\;
                                    \text{Outupt Dim}\;\Gamma_\text{Emb\_3} \\
                                    \text{Window Size}\;\Gamma_\text{WinSiz\_2},\;
                                    \text{Attention Heads}\;\Gamma_\text{HExp}
                                \end{bmatrix}\times 1$
                                }\\
    \cmidrule{2-4}
    &Stage 3 & Multi-Scale Block~\cite{wu2022tinyvit} &\makecell{$\begin{bmatrix}
                                    \text{Inupt Dim}\;\Gamma_\text{Emb\_3},\;
                                    \text{Outupt Dim}\;\Gamma_\text{Emb\_3} \\
                                    \text{Window Size}\;\Gamma_\text{WinSiz\_3},\;
                                    \text{Attention Heads}\;\Gamma_\text{HExp}^{2}
                                \end{bmatrix}\times 4$ \\
                                $\begin{bmatrix}
                                    \text{Inupt Dim}\;\Gamma_\text{Emb\_3},\;
                                    \text{Outupt Dim}\;\Gamma_\text{Emb\_4} \\
                                    \text{Window Size}\;\Gamma_\text{WinSiz\_3},\;
                                    \text{Attention Heads}\;\Gamma_\text{HExp}^{2}
                                \end{bmatrix}\times 1$
                                }\\
    \cmidrule{2-4}
    &Stage 4 & Multi-Scale Block~\cite{wu2022tinyvit} &\makecell{$\begin{bmatrix}
                                    \text{Inupt Dim}\;\Gamma_\text{Emb\_4},\;
                                    \text{Outupt Dim}\;\Gamma_\text{Emb\_4} \\
                                    \text{Window Size}\;\Gamma_\text{WinSiz\_4},\;
                                    \text{Attention Heads}\;\Gamma_\text{HExp}^{3}
                                \end{bmatrix}\times 2$ \\
                                }\\
    \bottomrule
    \end{tabular}
    }
  \label{tab:MobileSAM2Architecture}
\end{table*}

\subsection{Implementation Details}
We use approximately 10\% of the SA-V dataset ($\sim$ 11K videos) as our training data for efficient distillation. 
Following SAM2, we adopt the AdamW optimizer with an initial learning rate of $5\times 10^{-4}$. 
All models are trained for 5 epochs on an A100 (80GB) GPU with a batch size of 5 and 8-frame sequences per sample. 
Each training iteration uses a $4\times 4$ grid of spatial prompts for mask prediction by the teacher model, and we set the weighting coefficients $\alpha=1$ and $beta=1$ to balance temporal consistency and hierarchical knowledge transfer. 
The total training time is approximately 65 hours. 

In our progressive model contraction framework, we start with a hierarchical ViT seed (Hiera‑B) and progressively shrink one architectural axis at a time - embedding dimension ($\Gamma_\text{Emb}$), 
number of blocks ($\Gamma_\text{Blk}$), window size ($\Gamma_\text{WinSiz}$), and head expansion ($\Gamma_\text{HExp}$). 
Each reduction step produces lighter candidate models, 
all trained with the same distillation recipe, and evaluated on both accuracy and parameter count. The best candidate under a parameter budget is selected after each iteration, repeating this process until the desired efficiency or diminishing returns are observed. 
This approach avoids expensive large-scale searches, explicitly linking architectural choices to HyperKD and downstream validation rather than merely model size. 
Our lightweight architecture search resulted in approximately 15 candidate models; each candidate was quickly evaluated via a proxy: the image encoder was trained for a single epoch on roughly 10\% of SA-V, 
and candidates were selected based on accuracy measured on LVOS - each proxy run required $\sim$8 hours. 
LVOS was intentionally chosen, as our objective is long-term temporal consistency (a core strength of SAM2), 
and LVOS serves as the most relevant benchmark for this goal.

\subsection{Evaluation Metrics}

We use two widely adopted evaluation metrics, region similarity (J) and contour accuracy (F), following the DAVIS benchmark~\cite{perazzi2016benchmark}.

\begin{itemize}
    \item \textbf{Region Similarity (J)}: This metric calculates the Intersection over Union (IoU) between the ground truth (G) and the prediction (M), defined as:
    \begin{equation}
        \text{J} = \frac{\text{M} \cap \text{G}}{\text{M} \cup \text{G}}.
    \end{equation}
    
    \item \textbf{Contour Accuracy (F)}: This metric evaluates the precision of the segmentation boundary by computing the harmonic mean of contour recall ($\text{P}_\text{c}$) and contour precision ($\text{R}_\text{c}$), defined as:
    \begin{equation}
        \text{F} = \frac{2 \cdot \text{P}_\text{c} \cdot \text{R}_\text{c}}{\text{P}_\text{c} + \text{R}_\text{c}}.
    \end{equation}
\end{itemize}

Following the YouTube-VOS benchmark~\cite{xu2018youtube}, we separately report performance on seen and unseen categories. We calculate the final J\&F score as follows:
\begin{equation}
    \text{J\&F} = \frac{\text{J}_\text{s} + \text{F}_\text{s} + \text{J}_\text{u} + \text{F}_\text{u}}{4},
\end{equation}
where subscripts s and u denote scores for seen and unseen categories, respectively. This separate evaluation for seen and unseen categories provides a clearer measure of the model’s generalization ability in video object segmentation.

\subsection{Additional Discussions}
\subsubsection{Parameter Studies}

To investigate the influence of different loss function components in HyperKD, we perform a parameter study on the weighting coefficients $\alpha$ and $\beta$ in loss function Eq. (9).
We adjust $\alpha$ and $\beta$ over the range 
$[10^{-2},10^{-1},1,10^{1},10^{2}]$, 
while keeping the other parameter fixed at 1. 
The results are summarized in Table~\ref{tab:alpha_parameter_study} and Table~\ref{tab:beta_parameter_study}.

In Table~\ref{tab:alpha_parameter_study}, increasing $\alpha$ from $10^{-2}$ to $1$ significantly improves performance, indicating that the enforcement of temporal consistency is crucial for video segmentation.
At $\alpha=1$, the model achieves the best J\&F score of $69.0$.
For larger $\alpha$ values ($\alpha=10$ or $\alpha=10^2$), performance degrades because an excessive emphasis on temporal consistency might distort the boundaries of objects.

\begin{table}[h]
    \caption{Parameter Study on $\alpha$ in loss function Eq. (9). We adjust $\alpha$ while keeping $\beta=1$. The best performance is achieved when $\alpha=1$.}
    \label{tab:alpha_parameter_study}
    \centering
    \begin{tabular}{l|ccccc}
        \toprule
        Model Architecture & \multicolumn{5}{c}{MobileSAM2-23M} \\
        \midrule
        $\alpha$ & $10^{-2}$ & $10^{-1}$ & $1$ & $10^{1}$ & $10^{2}$ \\
        \midrule
        J\&F & 60.2 & 65.7 & \textbf{69.0} & 66.8 & 61.4 \\
        \bottomrule
    \end{tabular}
\end{table}

In Table~\ref{tab:beta_parameter_study}, similar to $\alpha$, performance improves as $\beta$ increases to 1, reaching a peak of $69.0$.
For very small $\beta$ values ($\beta=10^{-2}$ or $\beta=10^{-1}$), the model lacks sufficient hierarchical feature guidance, leading to lower accuracy.
For large $\beta$ values ($\beta=10^{1}$ or $\beta=10{2}$), performance drops, possibly due to overregularization from hierarchical distillation.

\begin{table}[h]
    \caption{Parameter Study on $\beta$ in loss function Eq. (9). We adjust $\beta$ while keeping $\alpha=1$. The best performance is achieved when $\beta=1$.}
    \label{tab:beta_parameter_study}
    \centering
    \begin{tabular}{l|ccccc}
        \toprule
        Model Architecture & \multicolumn{5}{c}{MobileSAM2-23M} \\
        \midrule
        $\beta$ & $10^{-2}$ & $10^{-1}$ & $1$ & $10^{1}$ & $10^{2}$ \\
        \midrule
        J\&F & 59.8 & 64.1 & \textbf{69.0} & 67.2 & 62.0 \\
        \bottomrule
    \end{tabular}
\end{table}

\subsubsection{Comparison with Different Hypergraph Construction Methods}
To investigate the impact of hypergraph formulation on knowledge distillation, we compare Fuzzy Hypergraph~\cite{mordeson2012fuzzy} and our proposed HyperKD-based hypergraph construction. The results are shown in Table~\ref{tab:hypergraph_study}.
Without any hypergraph-based knowledge distillation, the MobileSAM2-23M model achieves only 44.8 J\&F.
This shows that direct feature distillation without hypergraph modeling is insufficient for effective lightweight segmentation.
The Fuzzy Hypergraph~\cite{mordeson2012fuzzy} method improves the performance to 62.4 J\&F, indicating that using hypergraphs to model multi-way relationships enhances knowledge transfer.
However, Fuzzy Hypergraph~\cite{mordeson2012fuzzy} relies on predefined edge weights and lacks explicit high-order structure, which limits its ability to capture complex relationships in video segmentation.
HyperKD achieves the highest J\&F score of 69.0, significantly outperforming Fuzzy Hypergraph by +6.6.
This indicates that HyperKD's hypergraph formulation effectively captures high-order relations, leading to better temporal and granularity-aware knowledge transfer.

\begin{table}[h]
    \centering
    \caption{Comparison of Different Hypergraph Construction Methods. We evaluate the impact of different hypergraph formulations on the LVOS validation set (J\&F). HyperKD outperforms Fuzzy Hypergraph.}
    \label{tab:hypergraph_study}
    \begin{tabular}{l|c|cc}
        \toprule
        Model Architecture & \multicolumn{3}{c}{MobileSAM2-23M} \\
        \midrule
        Hypergraph Construction Method & No Distill & Fuzzy Hypergraph~\cite{mordeson2012fuzzy} & HyperKD (Ours) \\
        \midrule
        J\&F & 44.8 & 62.4 & \textbf{69.0} \\
        \bottomrule
    \end{tabular}
\end{table}

\subsubsection{Comparison to non-SAM2 Architecture Baseline in Semi-supervised VOS}
We conduct experiments to compare MobileSAM2 with MobileVOS shown in Table below.
Under the same 11K SA-V dataset training and semi-supervised VOS protocol, MobileSAM2 surpasses MobileVOS~\cite{Miles_2023_CVPR} with a stronger accuracy-size trade-off, enabled by HyperKD and architectural refinements that transfer SAM2’s temporal and multi‑granularity knowledge with minimal loss, 
validating MobileSAM2’s effectiveness and novelty.

\begin{table}[h]
    \centering
    \caption{Comparison to non-SAM2 architecture baseline in semi-supervised VOS. For fair comparsion, all methods process video sequences..}
    \resizebox{\linewidth}{!}{
\begin{tabular}	{ccc ccccc } 
\toprule
\multirow{2}*{{Model}}  &\multirow{2}*{{Distillation Method}} &\multirow{2}*{Params. (M)} & \multicolumn{5}{c}{J\&F}\\
& & & MOSE val & DAVIS2017 val &LOVS val &SA-V val &SA-V test \\
\midrule
\multicolumn{8}{c}{$\sim$ 10\% Training Data with 1 A100 GPU: 11K Videos from SA-V manual data}\\
\midrule
MobileNetV2 w/o ASPP & MobileVOS & 1.9 & 35.6 & 40.7 & 36.8 & 37.4 & 37.1 \\
MobileNetV2          & MobileVOS & 2.5 & 36.2 & 41.3 & 37.5 & 38.1 & 38.0 \\
ResNet18             & MobileVOS & 8.1 & 37.8 & 42.5 & 41.2 & 40.6 & 41.0 \\
\rowcolor{gray!16}MobileSAM2-5M & HyperKD (Ours) & 5.8 & 44.6 & 51.0 & 48.8 & 49.2 & 49.4 \\
\rowcolor{gray!16}MobileSAM2-10M & HyperKD (Ours) & 10.4 & 48.7 & 54.2 & 51.8 & 49.9 & 50.0 \\
\rowcolor{gray!16}MobileSAM2-23M & HyperKD (Ours) & 23.7 & 65.8 & 72.1 & 69.0 & 66.4 & 67.8 \\
\bottomrule
\end{tabular}
}
\end{table}

\subsubsection{Comparison to Recent Distillation Baseline}

We compare HyperKD to EA-KD~\cite{Su_2025_ICCV} under identical datasets/protocols below, indicating more effective transfer of temporal and multi-granularity knowledge. 

\begin{table}[h]
    \centering
    \caption{Comparison to Recent Distillation Baselines. We evaluate the impact of different hypergraph formulations on the LVOS validation set (J\&F).}
    \begin{tabular}	{c| c|cc } 
    \toprule
    Model Architecture & \multicolumn{3}{c}{MobileSAM2-23M} \\
    \midrule
    Distillation Method & No Distill &EA-KD~\cite{Su_2025_ICCV} & HyperKD (Ours)\\
    \midrule
    LOVS ({J\&F}) &44.8 &63.6 &\cellcolor{gray!16}{69.0} \\
    \bottomrule
    \end{tabular}
\end{table}

\subsubsection{Analysis of Data Proportion in HyperKD}

Table~\ref{tab:mobilesam2_data_sensitivity} shows the impact of the proportion of training data on HyperKD. 
J\&F scores increase from 45.2\% (1\% data) to 73.0\% (20\% data), with the largest gain between 5\% and 10\% (+13.2\%). 
In particular, with 10\% data, MobileSAM2-23M achieves 69.0\% J\&F, reaching 94.5\% of the performance at 20\%, demonstrating HyperKD’s data efficiency. 
Beyond 15\%, gains diminish (+1.7\%), indicating that HyperKD effectively extracts core knowledge with moderate data, reducing reliance on large-scale datasets. 
These results show that HyperKD enables near-optimal performance with only 10-15\% of data, making it ideal for low-resource training, real-time applications, and mobile deployment.

\begin{table}[h]
    \caption{Impact of Training Data Proportion on HyperKD Performance. We report J\&F scores (higher is better) on the LVOS validation set. HyperKD achieves competitive performance even with limited data.}
    \label{tab:mobilesam2_data_sensitivity}
    \centering
    \begin{tabular}{l|ccccc}
        \toprule
        Model Architecture & \multicolumn{5}{c}{MobileSAM2-23M} \\
        \midrule
        Training Data Proportion & 1\% & 5\% & 10\% & 15\% & 20\% \\
        \midrule
        J\&F & 45.2 & 55.8 & 69.0 & 71.3 & 73.0 \\
        \bottomrule
    \end{tabular}
\end{table}

\subsubsection{Comparison with lightweight non-distillation models}
As shown in Table~\ref{tab:mobilesam2_compare_with_non_distillation_model}, we conduct experiments to compare MobileSAM2 with FastSAM~\cite{zhao2023fast} and MobileSAM~\cite{zhang2023faster}, using the same evaluation datasets and evaluation protocols. As shown in Table below, fastSAM with 68M parameters, achieves a J\&F score of 48.9\%. MobileSAM with 5.78M parameters attains a considerably lower J\&F score of 32.2\%. In contrast, our MobileSAM2 achieves a J\&F score of 48.8\%, 51.8\%, and 69.0\% with only 5.84M, 10.37M, and 23.74M parameters, respectively, achieving a much better trade-off between model size and performance. The superior performance of MobileSAM2 largely attributed to our carefully designed HyperKD and architectural improvements, which effectively model and transfer SAM2’s rich temporal knowledge and diverse multi-granularity knowledge without excessive loss of accuracy. These results validate both the effectiveness and novelty of MobileSAM2 for lightweight segmentation.

\begin{table}[h]
    \caption{Comparison with lightweight non-distillation models. We report Parameters (MB) and J\&F scores (higher is better) on the LVOS validation set.}
    \label{tab:mobilesam2_compare_with_non_distillation_model}
    \centering
    \begin{tabular}{l|cc}
        \toprule
        Model Architecture & Params (MB)	 & J\&F \\
        \midrule
        FastSAM~\cite{zhao2023fast}	&68	&48.9\\
        MobileSAM~\cite{zhang2023faster}	&5.78	&32.2 \\
        \midrule
        MobileSAM2‑5M	&5.84	&48.8\\
        MobileSAM2‑10M	&10.37	&51.8\\
        MobileSAM2‑23M	&23.74	&69.0\\
        \bottomrule
    \end{tabular}
\end{table}

\subsubsection{Analysis of task-specific segmentation loss}
We evaluated HyperKD both with and without task-specific segmentation losses (i.e., cross-entropy using available masks) to assess their impact on optimization stability and performance. 
As shown in Table~\ref{tab:task-specific segmentation loss}, removing task-specific losses improves performance by +1.8 J\&F. We attribute this to the fact that task losses, derived from a limited 10\% labeled dataset, may introduce noise and conflict with the rich structural knowledge captured by the teacher. 
In contrast, HyperKD’s hypergraph-based supervision provides multi-level and high-quality guidance (temporal and granularity) that stabilizes optimization and reduces the need for additional task-specific objectives.

\begin{table}[h]
    \centering
    
    \caption{Analysis of task-specific segmentation loss. We evaluate the impact of using task-specific segmentation loss on the LVOS validation set (J\&F).}
    \label{tab:task-specific segmentation loss}
    \resizebox{\linewidth}{!}{
    \begin{tabular}{l|c|cc}
        \toprule
        Model Architecture & \multicolumn{3}{c}{MobileSAM2-23M} \\
        \midrule
        Distillation Method & No Distill & HyperKD (w/ task-specific loss)	& HyperKD (w/o task-specific loss) \\
        \midrule
        J\&F & 44.8 & 67.2 & \textbf{69.0} \\
        \bottomrule
    \end{tabular}
    }
\end{table}

\subsubsection{Study Limitations}
The progressive contraction strategy used to derive lightweight student models involves manually designed search spaces (e.g., embedding dimensions, number of blocks, attention head sizes). 
It lacks the automation and optimality guarantees of neural architecture search, potentially limiting the performance ceiling of MobileSAM2.

\subsubsection{Visualization of High-Order Learning in HyperKD}
\begin{figure}[t]
    \centering
    \includegraphics[width=0.75\linewidth]{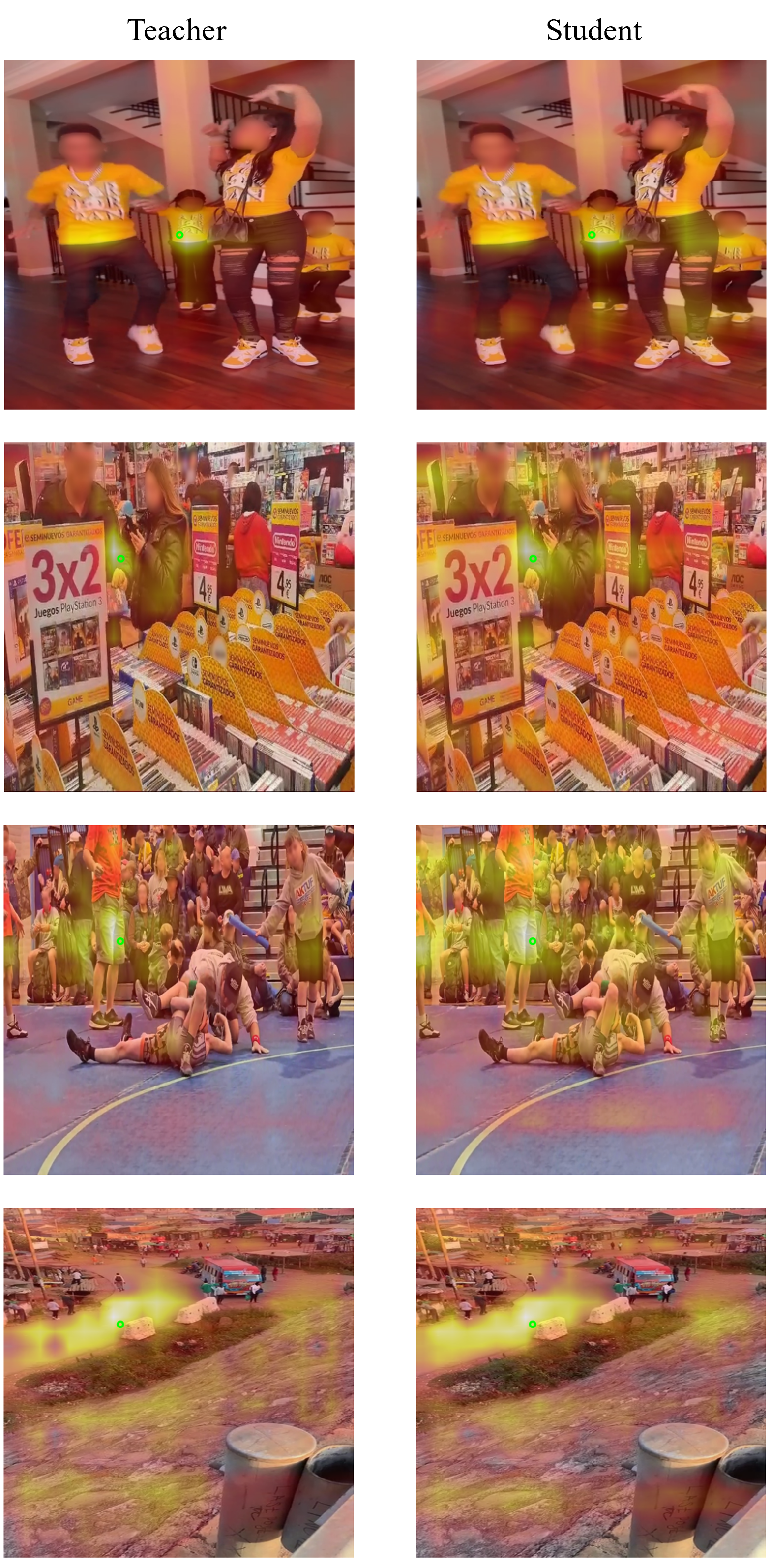}
    \caption{Visualization of High-Order Learning in HyperKD. }
    \label{fig:query_sim}
\end{figure}

Fig.~\ref{fig:query_sim} visualizes the hypergraph affinity structure via query-pixel similarity. 
For a selected query point on an object (marked in green), 
we compute the cosine similarity in the teacher's feature space between that point and every other spatial position, producing a heatmap that reveals which pixels the hypergraph considers semantically connected. 
The teacher's hypergraph cleanly groups the queried object, with high similarity concentrated on the object's full extent and minimal leakage to the background. 
The student distilled via HyperKD produces a closely matching similarity map, confirming that the core hypergraph structure—pixel-level semantic grouping—is successfully transferred from the 224M teacher to the 23M student.

\subsection{Qualitative Results}
\begin{figure}[t]
    \centering
    \includegraphics[width=1\linewidth]{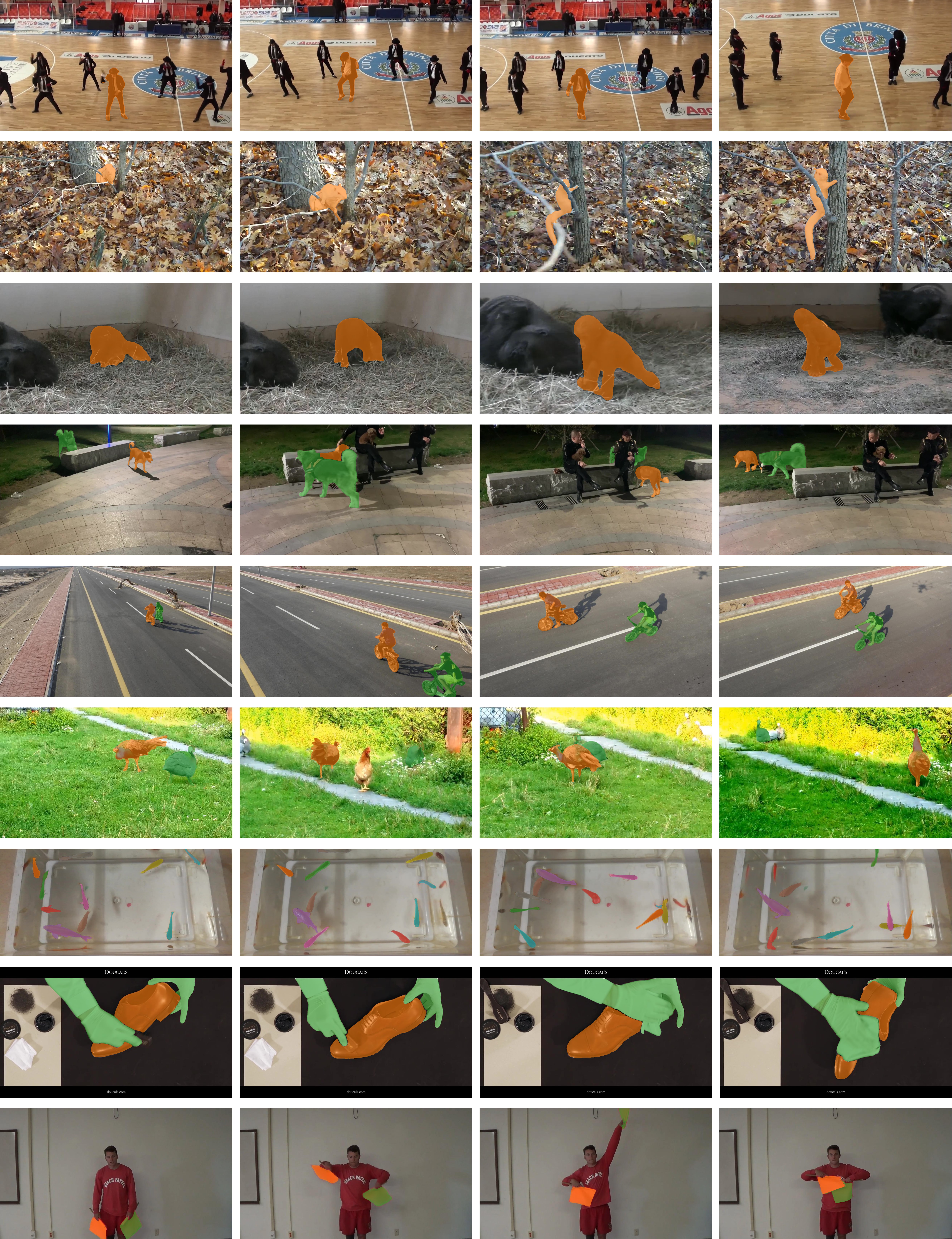}
    \caption{Qualitative illustrations of MobileSAM2-24M, where each row refers to frames from one video and each masklet has one unique color. }
    \label{fig:result_vis}
\end{figure}

We provide more qualitative illustrations of our MobileSAM2-24M.
As shown in Figure \Cref{fig:result_vis}, MobileSAM2 produces segmentation across multiple video consistently.

\end{document}